\documentclass[final,5p,times,twocolumn]{elsarticle}

\usepackage{amsmath,amssymb}
\usepackage{graphicx}
\usepackage{booktabs}
\usepackage{tabularx}
\usepackage{array}
\usepackage{multirow}
\usepackage{xcolor}
\usepackage{microtype}
\usepackage{colortbl}
\usepackage[hyphens]{url}
\usepackage{enumitem}
\usepackage[colorlinks=true,linkcolor=blue!60!black,
            citecolor=blue!60!black,urlcolor=blue!60!black,
            pdftitle={Trustworthy Agentic AI: Failure Modes, Mitigation
              Strategies, and a Lifecycle Framework for Autonomous LLM Systems},
            pdfauthor={Fayeq Jeelani Syed, Rehan Ahmed,
              Ali Albatayineh, Ahmad Javaid},
            pdfkeywords={agentic AI, LLM agents, AI safety, alignment,
              trustworthy AI, autonomous agents, AI governance, TADL}]{hyperref}
\usepackage{cleveref}   
\usepackage{placeins}  

\newcolumntype{L}[1]{>{\raggedright\arraybackslash}p{#1}}
\newcolumntype{C}[1]{>{\centering\arraybackslash}p{#1}}

\definecolor{rowA}{RGB}{242,242,242}

\usepackage{tikz}
\usetikzlibrary{shapes.geometric,arrows.meta,positioning,fit,backgrounds,calc}

\tikzset{
  phase/.style={rectangle,rounded corners=4pt,draw=blue!50!black,
                fill=blue!12,text width=2.0cm,align=center,
                minimum height=0.8cm,font=\small\bfseries},
  arrow/.style={-{Stealth[length=5pt]},thick,blue!60!black},
  gate/.style={diamond,draw=orange!70!black,fill=orange!15,
               minimum size=0.55cm,inner sep=1pt,font=\scriptsize},
  artifact/.style={rectangle,draw=gray!60,fill=gray!10,
                   text width=1.6cm,align=center,font=\scriptsize,
                   minimum height=0.6cm,rounded corners=2pt},
}

\begin{document}
\sloppy

\begin{frontmatter}

\title{Trustworthy Agentic AI: Failure Modes, Mitigation Strategies, and a Lifecycle Framework for Autonomous LLM Systems}

\author[IUI]{Fayeq Jeelani Syed\corref{cor1}}
\ead{sjeelan@iu.edu}
\cortext[cor1]{Corresponding author.}

\author[Purdue]{Rehan Ahmed}
\ead{ahmad108@pnw.edu}

\author[HCT]{Ali Al Bataineh}
\ead{aalbataineh@hct.ac.ae}

\author[Yale]{Aakriti Adhikari}
\ead{aakriti.adhikari@yale.edu}

\affiliation[IUI]{%
  organization={Department of Human-Centered Computing,
  Indiana University Indianapolis},
  addressline={755 W. Michigan Street},
  city={Indianapolis},
  state={Indiana},
  postcode={46202},
  country={USA}
}

\affiliation[Purdue]{%
  organization={Department of Electrical and Computer Engineering,
  Purdue University Northwest},
  addressline={2200 169th Street},
  city={Hammond},
  state={Indiana},
  postcode={46323},
  country={USA}
}

\affiliation[HCT]{%
  organization={Computer Information Science,
  Higher Colleges of Technology},
  city={Abu Dhabi},
  country={United Arab Emirates}
}

\affiliation[Yale]{%
  organization={Department of Biomedical Informatics and Data Science,
  Yale School of Medicine, Yale University},
  addressline={333 Cedar Street},
  city={New Haven},
  state={Connecticut},
  postcode={06510},
  country={USA}
}

\begin{abstract}
Agentic AI systems built on large language models can plan over multiple steps, use external tools, retain information in memory, and coordinate with other agents. These capabilities make them more useful than static language models, but they also introduce new security and operational risks. Untrusted content from websites, emails, documents, and databases can enter the same context as system instructions; persistent memory can carry compromised information across sessions; and access to external tools can turn an incorrect model response into a consequential real-world action. This article reviews the trustworthiness of agentic AI across five interconnected dimensions: safety and robustness, alignment and human oversight, transparency and auditability, privacy and data governance, and regulatory compliance. It organizes key failure modes—including indirect prompt injection, backdoor triggers, goal misgeneralization, memory contamination, and cross-session data leakage—into a unified taxonomy. It also examines major mitigation approaches, such as instruction hierarchies, context isolation, spotlighting, process-based supervision, constrained tool use, and privacy-preserving memory, while distinguishing techniques supported by empirical evidence from those that remain largely conceptual. Building on this analysis, we introduce the Trustworthy Agent Development Lifecycle (TADL), a six-phase framework covering specification, design, training, evaluation, deployment, and monitoring. For each phase, TADL identifies relevant trust activities, expected evidence, and risk-based decision gates. Although TADL has not yet been empirically validated, it provides a structured foundation for developing and evaluating more secure and accountable agentic systems. The article concludes by identifying gaps in current benchmarks and outlining priorities for future research.

\end{abstract}


\end{frontmatter}

\section{Introduction}
\label{sec:intro}

Artificial intelligence has moved through several broad phases,
from symbolic rules to statistical learning, deep networks, and
large pre-trained foundation models. The latest change is from a
language model that responds to a prompt to an \emph{agent} that
can decide what to do next. LLM-based agents may plan over many
steps, use web browsers, code interpreters, databases, or APIs,
communicate with other agents, and take actions that affect the
outside world~\citep{yao2023react,wu2024autogen}.

That change matters for risk. A false statement from a chatbot can
usually be checked and corrected. An agent acting in a financial
account, clinical workflow, or infrastructure system may turn the
same underlying model error into an action that is difficult to
reverse. It may also repeat or compound the error before a person
notices. Research on these problems is growing, but it is still
divided across safety, alignment, privacy, explainability, and
governance communities. Technical defences are not always
considered alongside deployment controls, while policy proposals
do not always reflect how agents operate in practice.

This paper addresses that separation by examining the major
trustworthiness questions together rather than as independent
problems.
Several recent surveys address adjacent concerns---LLM safety
and agent security~\citep{dong2024safeguarding,yu2025trustworthy},
agent architectures~\citep{abouali2025agentic}, or specific
threats such as prompt injection~\citep{greshake2023not}---but
relatively little work connects all five trust dimensions to an
operational lifecycle framework.
\Cref{sec:related} characterises how this survey relates to and
complements that body of work.

\subsection{Scope and Contributions}
\label{sec:contributions}

This survey makes the following contributions:

\begin{itemize}[leftmargin=*, itemsep=2pt]
  \item A structured \textbf{taxonomy of agentic failure modes}
    organised across five trust dimensions: safety, alignment,
    transparency, privacy, and governance
    (\Cref{sec:taxonomy}).

  \item A \textbf{structured narrative review} of representative,
    influential, and directly relevant literature across those
    dimensions, covering publications primarily from 2018 to
    mid-2026 (Sections~\ref{sec:safety}--\ref{sec:governance}).

  \item A \textbf{comparative discussion} of principal mitigation
    families, their trade-offs, evidence base, and deployment
    maturity (Sections~\ref{sec:safety}--\ref{sec:governance}).

  \item The \textbf{Trustworthy Agent Development Lifecycle
    (TADL)}, an author-proposed conceptual six-phase framework
    that organises trust requirements across the agent development
    and deployment lifecycle (\Cref{sec:tadl}).

  \item Identification of \textbf{evaluation and benchmarking
    gaps} and \textbf{open research problems}
    (\Crefrange{sec:evalgaps}{sec:openproblems}).
\end{itemize}

\noindent
Most existing surveys concentrate on one part of the problem, such
as agent architecture, security threats, controllability,
evaluation, or trust and risk management. Here, those strands are
considered together. The discussion links safety, alignment,
transparency, privacy, and governance to TADL, an operational
lifecycle framework with phase-specific controls, evidence
artefacts, trust gates, monitoring, and escalation procedures.

\subsection{Why Agentic AI Demands a New Trustworthiness Framework}
\label{sec:why}

Five features distinguish agentic systems from conventional ML
deployments and help explain why their risk profile is different.

\begin{enumerate}[leftmargin=*, itemsep=2pt]
  \item \textbf{Temporal extent.} Agents act over sequences of
    steps; errors compound and drift from intended goals in ways
    that single-inference models cannot.
  \item \textbf{Environmental reach.} Tool use connects agents to
    external systems---code execution, email, APIs---dramatically
    expanding the attack surface and consequence space.
  \item \textbf{Reduced oversight density.} Agents operate between
    human checkpoints; misalignment or adversarial manipulation
    may persist across many steps undetected.
  \item \textbf{Multi-agent dynamics.} Interacting agents exhibit
    emergent behaviours not predictable from individual properties,
    creating collective alignment challenges.
  \item \textbf{Open-endedness.} Agents encounter situations their
    designers did not anticipate, requiring safety properties to
    generalise beyond the training distribution.
\end{enumerate}

Techniques used with single-turn LLMs, including RLHF, output
filtering, and content moderation, still matter. On their own,
however, they do not address this broader risk profile.

\subsection{Literature Search Methodology}
\label{sec:methodology}

We conducted a \textbf{structured narrative review}; this was not
a PRISMA-style systematic review. Relevant literature was located
through searches of arXiv (cs.AI,
cs.CL, cs.CR, cs.LG); proceedings of NeurIPS, ICML, ICLR, ACL,
EMNLP, IEEE S\&P, and ACM CCS; Springer, Elsevier, and ACM
Digital Library; and official government and standards-body
websites for regulatory documents.

Search terms included \emph{agentic AI},
\emph{LLM agents}, \emph{autonomous agents}, \emph{AI safety},
\emph{alignment}, \emph{trustworthy AI}, \emph{prompt injection},
\emph{AI governance}, and \emph{multi-agent systems}.
The review covers work published roughly between
2018 and mid-2026, with greater attention to the period after 2022
when research on LLM-based agents expanded quickly. Papers from
2025 and 2026 were selected for their relevance rather than their
citation counts, which have not had time to stabilise.

Because the sources cited here carry different kinds of
authority, we treat four categories differently rather than
applying a single inclusion rule.
\emph{Peer-reviewed publications} are the default basis for
technical claims about what a method does or how well it
performs.
\emph{Preprints} are cited where they represent a significant
contribution for which no peer-reviewed version could be located
at the time of writing, or where the preprint remains the main
citable version of an influential contribution; such claims are
attributed to the authors rather than stated as established
results, and the reliance on preprints is noted as a limitation
in \Cref{sec:limitations}.
\emph{Standards and legislation} are cited from the issuing body
or the official legal gazette rather than from secondary
commentary, and are used to establish what an instrument requires,
not whether it is effective.
\emph{Official policy documents} are cited to describe stated
government positions and are not treated as evidence about
outcomes.
Where a preprint has since appeared in a peer-reviewed venue, the
published version is cited.

The result is representative rather than exhaustive, and no claim
of complete coverage is made. Work published after mid-2026 falls
outside the review, and recent or cross-disciplinary studies may
have been missed---particularly at the boundary between
regulatory scholarship and technical AI research. Direct
comparison across studies is also difficult because agent
evaluations use different tasks, threat models, and reporting
practices.

\begin{table*}[!htbp]
\centering
\caption{Comparison of the present survey with closely related
         prior surveys on key trustworthiness dimensions.
         Coverage levels: S = Substantial (primary focus with
         analysis and mitigations); P = Partial (discussed but
         not the primary focus); L = Limited (mentioned or briefly
         noted); -- = Not a primary focus.
         For all surveys including this one, ratings are author
         assessments of the \emph{depth of treatment} a dimension
         actually receives, not an independent quality assessment.
         The same rubric was applied to the present survey; where a
         dimension is surveyed to identify gaps rather than analysed
         in depth, it is rated P rather than S.
         Reasonable disagreement is possible; readers are
         encouraged to consult the original works.}
         
\label{tab:surveycomparison}
\renewcommand{\arraystretch}{1.15}
{\setlength{\tabcolsep}{3pt}
\scriptsize
\begin{tabularx}{\textwidth}{
  @{}
  >{\raggedright\arraybackslash}p{3.0cm}
  *{8}{>{\centering\arraybackslash}X}
  @{}
}
\toprule
\textbf{Survey} &
\textbf{Arch.} &
\shortstack{\textbf{Safety/}\\\textbf{Security}} &
\shortstack{\textbf{Alignment/}\\\textbf{Oversight}} &
\shortstack{\textbf{Transparency}} &
\shortstack{\textbf{Privacy}} &
\shortstack{\textbf{Governance}} &
\shortstack{\textbf{Life cycle}} &
\shortstack{\textbf{Agent}\\\textbf{Evaluation}} \\
\midrule

\rowcolor{rowA}
Yu et al.\ (2025)~\citep{yu2025trustworthy} &
P & S & P & P & S & L & -- & P \\

Raza et al.\ (2026)~\citep{raza2025trism} &
P & S & P & P & P & S & P & P \\

\rowcolor{rowA}
Abou Ali et al.\ (2026)~\citep{abouali2025agentic} &
S & P & L & L & L & P & -- & P \\

Gan et al.\ (2024)~\citep{gan2024navigating} &
P & S & L & P & S & L & -- & L \\

\rowcolor{rowA}
\textbf{This survey} &
P & S & S & S & S & S & S & P \\

\bottomrule
\end{tabularx}
}
\end{table*}

\subsection{Related Surveys and Positioning}
\label{sec:related}

Several recent surveys cover parts of the trustworthy-agent
problem. The closest ones differ mainly in what they treat as the
unit of analysis and how far they extend from technical threats
into deployment and governance.

\citet{yu2025trustworthy} provide a survey on trustworthy LLM
agents focusing on threats (adversarial attacks, jailbreaks,
privacy leaks) and corresponding countermeasures, organised
around agent modules.
Its attack taxonomy is more detailed than ours. Our emphasis is
instead on connecting those attacks with alignment, governance,
and controls applied across the system lifecycle.

\citet{raza2025trism} survey trust, risk, and security management
in LLM-based agentic multi-agent systems through the lens of the
TRiSM (Trust, Risk, and Security Management) framework,
emphasising risk assessment and security governance.
Their analysis centres on security-management processes. We cover
alignment oversight, transparency, privacy, and regulatory
compliance in addition, and use a lifecycle framework to connect
them.

\citet{abouali2025agentic} provide a comprehensive architectural
survey of agentic AI systems across symbolic and neural paradigms,
covering healthcare, finance, and robotics applications.
Their primary concern is architectural taxonomy and application
domains, whereas ours is the set of trust properties and
operational controls needed around those architectures.

\citet{gan2024navigating} survey security, privacy, and ethics
threats in LLM-based agents, identifying attack surfaces across
perception, memory, reasoning, and action modules.
We draw on their threat characterisation but place it within a
broader discussion of mitigation, regulation, and lifecycle
controls.

\Cref{tab:surveycomparison} compares the present survey with
the most directly relevant prior works on dimensions relevant to
trustworthiness coverage.

\subsection{Paper Organisation}
\label{sec:organisation}

\Cref{sec:background} introduces the agentic AI paradigm.
\Cref{sec:taxonomy} presents the failure-mode taxonomy.
Sections~\ref{sec:safety}--\ref{sec:governance} address the five
trust dimensions.
\Cref{sec:tadl} proposes TADL.
\Cref{sec:evalgaps} discusses evaluation and benchmarking gaps.
\Cref{sec:openproblems} identifies open research problems.
\Cref{sec:limitations} states survey limitations.
\Cref{sec:conclusion} concludes.

\section{Background: Agentic AI Architectures and Capabilities}
\label{sec:background}

\subsection{Defining Agentic AI}
\label{sec:definition}

The term ``agent'' predates LLMs by several
decades~\citep{wooldridge1995intelligent}. In this paper,
\emph{agentic AI} refers more narrowly to LLM-based systems that
combine language understanding and generation with an autonomous
agent's ability to plan and act. We use the following working
definition:

\begin{quote}
\textit{An agentic AI system is an LLM-based system that
(1)~receives a high-level goal from a human principal;
(2)~autonomously decomposes it into sub-tasks;
(3)~executes actions by invoking tools, other agents, or
external APIs; (4)~observes results and updates its plan; and
(5)~iterates until the goal is achieved or a stopping condition
is met, with human oversight at a frequency substantially lower
than the action frequency.}
\end{quote}

This encompasses frameworks such as ReAct~\citep{yao2023react},
AutoGen~\citep{wu2024autogen}, and LangChain-based
agents~\citep{chase2022langchain}.
It excludes single-turn LLM inference and chain-of-thought
prompting without external action.

\subsection{Core Architectural Components}
\label{sec:components}

Implementations vary, but most agentic systems contain some version
of the following components. Their interactions matter as much as
the individual modules because failures often cross module
boundaries.

\begin{itemize}[leftmargin=*, itemsep=2pt]
  \item \textbf{Perception module.} Ingests observations:
    user instructions, tool outputs, memory retrievals,
    inter-agent messages.
  \item \textbf{Planning and reasoning engine.} An LLM that
    interleaves reasoning with action selection and observations,
    as in ReAct~\citep{yao2023react}.
  \item \textbf{Memory system.} In-context state and episodic
    records, as illustrated by the memory stream in Generative
    Agents~\citep{park2023generative}; implementations may also
    use external retrieval stores.
  \item \textbf{Tool executor.} Invokes web search, code
    execution, file access, and APIs.
  \item \textbf{Output and action interface.} Formats and
    delivers outputs or actions downstream.
  \item \textbf{Orchestrator (multi-agent).} Assigns tasks to
    sub-agents and aggregates
    results~\citep{wu2024autogen}.
\end{itemize}

\subsection{Representative Architectures}
\label{sec:architectures}

\Cref{tab:architectures} summarises five broad architecture
classes with their distinctive trust challenges.
``Copilot paradigms'' and ``CI/CD agents'' are used here as
descriptive labels for documented deployment patterns rather than
specific named products.

\begin{table}[!htbp]
\centering
\caption{Agentic AI architecture classes and distinctive
         trustworthiness challenges.}
\label{tab:architectures}
\scriptsize
\renewcommand{\arraystretch}{1.15}
\begin{tabularx}{\columnwidth}{L{1.8cm}L{2.5cm}L{2.5cm}}
\toprule
\textbf{Architecture} & \textbf{Trust Challenges} &
\textbf{Examples} \\
\midrule
\rowcolor{rowA}
Single-Agent ReAct &
  Goal drift; tool misuse; context-length limits &
  ReAct~\citep{yao2023react} \\
Hierarchical Multi-Agent &
  Cascading failures; authority confusion &
  AutoGen~\citep{wu2024autogen}; MetaGPT~\citep{hong2023metagpt} \\
\rowcolor{rowA}
Collaborative Peer Agents &
  Convergence on an incorrect view; judge bias &
  LLM debate~\citep{liang2023encouraging} \\
Human-in-the-Loop (HITL) &
  Approval fatigue; automation bias &
  HITL agent deployment patterns \\
\rowcolor{rowA}
Fully Autonomous Pipeline &
  All failure modes amplified &
  Automated software-engineering agents \\
\bottomrule
\end{tabularx}
\end{table}

\subsection{Agent Benchmarks and the Capability--Trust Gap}
\label{sec:gap}

Several benchmarks now measure what agents can accomplish.
SWE-bench evaluates LLM-based resolution of real GitHub
issues~\citep{jimenez2024swebench}; WebArena tests web navigation
across realistic environments~\citep{zhou2024webarena}; and GAIA
measures general AI-assistant
capability~\citep{mialon2023gaia}.
AgentBench~\citep{liu2023agentbench} provides a broader
multi-task evaluation environment.
These benchmarks document capability on their respective task
sets; they do not by themselves measure adversarial robustness,
long-horizon alignment, or practical interpretability. We refer
to the resulting evaluation imbalance as the
\emph{capability--trust gap}.

The gap is not only theoretical. Indirect prompt injection has
redirected tool-using applications and exposed data in evaluated
attack settings~\citep{greshake2023not}. Separately, an erroneous
recommendation can become more consequential when an agent is
authorised to act on it rather than merely display it.

\subsection{Deployment Risk Tiers}
\label{sec:risktiers}

The safeguards required for a scheduling assistant should not be
the same as those required for a clinical agent. We therefore use
four risk tiers based on the likely consequences of failure:

\begin{itemize}[leftmargin=*, itemsep=2pt]
  \item \textbf{Critical} (irreversible harm): healthcare,
    critical infrastructure, legal systems.
  \item \textbf{High} (significant but recoverable): code
    deployment, HR pipelines, financial account management.
  \item \textbf{Moderate} (limited consequence): content
    generation, research assistance, scheduling.
  \item \textbf{Low} (easily verified): information retrieval,
    human-reviewed drafting.
\end{itemize}

Problems arise when safeguards suited to Low-tier uses are carried
into Critical- or High-tier deployments, often because deployment
is moving faster than risk assessment.

\section{Taxonomy of Agentic Failure Modes}
\label{sec:taxonomy}

Agent failures combine familiar LLM weaknesses with the dynamics
of autonomous, multi-step action. A hallucination, for example,
can become an incorrect tool call and then influence later steps.
We group these failures into the same five trust dimensions used
throughout the paper.
\Cref{tab:failuremodes} provides illustrative examples with
author-assigned illustrative severity ratings.

\textbf{Note on severity ratings.}
The ratings in \Cref{tab:failuremodes} are \emph{author-assigned
illustrative assessments}, not empirically validated risk scores.
They reflect qualitative judgements based on four criteria:
(1)~scope of possible harm (individual vs.\ systemic);
(2)~reversibility of harm;
(3)~degree of agent autonomy at the point of failure; and
(4)~detectability before harm occurs.
Where consequences depend heavily on deployment context, the
rating reflects a typical high-stakes scenario.

\subsection{Safety Failures}
Direct harm execution; cascading failures across connected
systems; resource exhaustion through runaway API or compute use;
unsafe tool invocation violating access constraints.

\subsection{Alignment Failures}
Goal misgeneralization---the agent pursues a proxy objective that
aligned during training but diverges at
deployment~\citep{shah2022goal};
specification gaming that satisfies a measured metric while
violating intent~\citep{shah2022goal};
instruction misinterpretation yielding technically compliant but
harmful outcomes; value drift over extended operation.

\subsection{Transparency Failures}
Opaque intermediate reasoning preventing meaningful oversight;
post-hoc rationalisation where stated reasoning does not reflect
the actual computational
process~\citep{turpin2023language,lanham2023measuring};
overconfident outputs increasing automation bias; deceptive
alignment (largely theoretical at present capability
levels~\citep{hubinger2019risks}).

\subsection{Privacy Failures}
Data leakage through over-permissioned tool calls; memory
contamination accumulating sensitive data across sessions;
extraction of memorised private information through model
outputs~\citep{carlini2021extracting}; and extraction of private
user--agent interactions from persistent agent
memory~\citep{wang2025privacy}.

\subsection{Governance Failures}
Compliance violations in regulated domains (HIPAA, GDPR, MiFID
II); accountability gaps when harm cannot be attributed across
the principal hierarchy; audit-trail deficiencies preventing
forensic review; cross-jurisdictional regulatory conflicts.

\begin{table}[!htbp]
\centering
\caption{Taxonomy of agentic failure modes with illustrative
         examples and author-assigned illustrative severity
         ratings (C = Critical; H = High). Ratings are based on
         scope of harm, reversibility, agent autonomy, and
         detectability, and reflect typical high-stakes
         scenarios; actual severity is context-dependent.}
\label{tab:failuremodes}
\renewcommand{\arraystretch}{1.1}
\scriptsize
\begin{tabularx}{\columnwidth}{L{1.3cm}L{1.9cm}L{2.7cm}C{0.6cm}}
\toprule
\textbf{Dimension} & \textbf{Sub-type} & \textbf{Example} &
\textbf{Illus.\ Sev.} \\
\midrule
\rowcolor{rowA}
Safety & Unsafe tool use &
  Agent executes attacker-injected code & C \\
Safety & Cascading failure &
  Cloud misconfiguration triggers data loss & C \\
\rowcolor{rowA}
Safety & Resource exhaustion &
  Loop causes 10{,}000 unintended API calls & H \\
Alignment & Goal misgeneralization &
  Agent spams to maximise ``email responses'' & C \\
\rowcolor{rowA}
Alignment & Spec.\ gaming &
  Agent marks task complete without solving it & H \\
Alignment & Instruction misinterp. &
  Agent deletes active files matching ``old'' & H \\
\rowcolor{rowA}
Transparency & Opaque reasoning &
  Loan decision not auditable & H \\
Transparency & Post-hoc rationalisation &
  Fabricated citations in medical report & C \\
\rowcolor{rowA}
Privacy & Data leakage &
  PII sent to third-party search API & H \\
Privacy & Memory contamination &
  Prior user's health data recalled in new session & C \\
\rowcolor{rowA}
Governance & Compliance violation &
  Investment advice without required disclosure & H \\
Governance & Accountability gap &
  No identifiable liable party after agent harm & H \\
\bottomrule
\end{tabularx}
\end{table}

\section{Safety and Robustness}
\label{sec:safety}

\subsection{Prompt Injection}
\label{sec:promptinjection}

Prompt injection takes advantage of a basic weakness in current
LLMs: they do not reliably separate trusted instructions from
untrusted text in their context~\citep{greshake2023not,perez2022ignore}.
The problem becomes more serious for agents because reading web
pages, emails, and database records is part of their normal work.
Any of these sources can contain text that the model mistakenly
treats as an instruction~\citep{yi2023prompt}.

\citet{greshake2023not} demonstrated indirect prompt injection
against several deployed LLM-integrated applications.
\citet{zhan2024injecagent} benchmarked indirect injection
against 30 tool-integrated agents using 1{,}054 test cases,
finding widespread vulnerability: a ReAct-prompted GPT-4 agent
was successfully attacked in roughly one case in four, with
success rates rising further when the injected instruction was
reinforced.
\citet{debenedetti2024agentdojo} introduced AgentDojo, a dynamic
benchmark for evaluating prompt injection attacks and defences.

The available defences intervene at different layers, and their
results should not be treated as directly comparable.
\emph{Instruction hierarchies}~\citep{wallace2024instruction}
train models to assign higher weight to positional or
token-marked trusted instructions; they reduce but do not
eliminate injection risk and were evaluated primarily on static
LLMs rather than deployed agents.
\emph{Context isolation} processes environmental data in separate
contexts with no write access to core instruction registers; it
is effective but constrains agents' ability to integrate
information across sources and adds system complexity.
\emph{Detection-based defences} evaluated in AgentDojo inspect
content for attacks~\citep{debenedetti2024agentdojo}; their measured
utility depends on the benchmark's attack distribution and on the
cost of blocking benign tasks.
\emph{Spotlighting}~\citep{hines2024defending} marks untrusted
data with special tokens to aid the model in distinguishing it
from instructions. Its evidence is strongest for controlled
prompt-injection tests; that result does not establish resilience
when an agent repeatedly retrieves and transforms hostile content.
\emph{Formal action constraints} restrict the action set
regardless of reasoning; they provide strong guarantees but
reduce agent flexibility.

These results point to an architectural problem rather than a flaw
that can be removed with a better system prompt. If trusted
instructions and untrusted environmental content occupy the same
context, some risk of instruction confusion remains. Spotlighting
changes the representation of untrusted text within that context;
it is evidence for source marking, not for complete architectural
isolation~\citep{hines2024defending}.
Even so, the studies use different agents, attack sets, and success
criteria. They establish partial risk reduction, not a general
solution to prompt injection.

\subsection{Goal Misgeneralization and Specification Gaming}
\label{sec:misgeneralization}

Goal misgeneralization occurs when a learned policy achieves
intended objectives within the training distribution but pursues
a different objective at
deployment~\citep{shah2022goal}.
Specification gaming involves the agent optimising an explicit
metric while violating the spirit of the
task~\citep{shah2022goal}.

With LLM agents, feedback-based fine-tuning may reward the
\emph{appearance of correctness} instead of successful completion
of the underlying task. Over a long trajectory, even a small
difference between the intended and learned objective can affect
many later actions before it is detected.

Process reward models (PRMs)~\citep{lightman2024lets} evaluate
intermediate reasoning steps, providing denser supervision that
can expose errors before the final answer; evidence comes primarily from mathematical
reasoning tasks. That evidence supports better supervision of
locally checkable steps, but does not show that PRMs detect an
agent pursuing the wrong objective across a long trajectory.
Constitutional AI (CAI)~\citep{bai2022constitutional} encodes
normative principles directly into training, reducing dependence
on narrow reward signals; its effectiveness for long-horizon
tasks has not been established by the cited experiments.
Debate-based evaluation~\citep{irving2018ai} leverages agent
self-critique. Unlike PRMs, debate attempts to make evaluation
easier without requiring a labelled score for every step; its
weakness is dependence on a human judge who may be persuaded by
a fluent but incorrect argument.
AgentDojo provides an adversarial evaluation environment for a
narrower question---whether prompt injection changes task outcomes
or causes security violations~\citep{debenedetti2024agentdojo}.

\subsection{Robustness to Distribution Shift}
\label{sec:robustness}

Distribution shift is especially difficult for agents because
their own actions can change the environment and create feedback
loops.
Tool API drift (external interfaces change over time), adversarial
environment evolution, world-model obsolescence in persistent
memory, and context-length extrapolation beyond training horizons
are the principal shift mechanisms.

Elastic weight consolidation was introduced to reduce
catastrophic forgetting in sequential learning
settings~\citep{kirkpatrick2017overcoming}. Applying that idea to
LLM-agent adaptation is an extrapolation: it does not address tool
API drift or guarantee safe behaviour after environmental change.
Uncertainty-triggered human escalation addresses a different
problem by limiting action when confidence is inadequate.

\subsection{Multi-Agent Safety}
\label{sec:multiagentsafety}

Multi-agent systems add problems that do not arise in quite the
same way for a single agent. Agents that
appear safe when tested separately may produce an unsafe result
when their decisions interact. AutoGen's conversation framework
illustrates how an orchestrator can route instructions and
aggregate subordinate outputs~\citep{wu2024autogen}; the claim that
compromising that role could affect the group is an architectural
inference, not a security result reported by the AutoGen paper.
Collective alignment---ensuring that a system of agents pursues
human-aligned goals---has no satisfactory solution at present.

Cryptographic attestation could establish who sent an inter-agent
message, but it cannot establish that the message is correct or
aligned. Formal verification offers a different guarantee---that
a specified protocol property holds---and inherits the familiar
difficulty of writing a complete specification. These approaches
therefore address different parts of the problem rather than
competing as interchangeable solutions.

\subsection{Formal Safety Constraints}
\label{sec:formalsafety}

Behavioural training is not the only way to limit harm. The action
space can also be restricted so that certain catastrophic actions
are unavailable regardless of what the model concludes.
Runtime enforcement of action invariants provides strong,
verifiable guarantees.
The primary challenge is \emph{specification completeness}:
anticipating every unsafe action class is difficult in
open-ended environments.
\citet{dalrymple2024towards} instead propose a broader guaranteed-
safety programme combining specifications, world models, and
verifiers. This is a research framework, not evidence that a
particular hybrid architecture is already practical for deployed
LLM agents.

Backdoors are a separate robustness concern. Experiments on web
shopping and tool-use tasks show that triggers placed in user
queries or intermediate environmental observations can manipulate
agent behaviour~\citep{yang2024watch}.

\section{Alignment and Human Oversight}
\label{sec:alignment}

\subsection{The Value Specification Problem}
\label{sec:valuespec}

Human values depend on context, differ across people and cultures,
and can conflict with one another; deciding \emph{whose} values an
agent should be aligned to is a normative question distinct from
the technical one of how to encode
them~\citep{gabriel2020artificial}. Current work approaches the
technical side mainly through three families of methods.

\textbf{RLHF}~\citep{christiano2017deep,ziegler2019fine} trains
a reward model on human preference comparisons and uses it to
fine-tune the agent via reinforcement learning.
RLHF has produced strong alignment results for single-turn LLMs
but faces amplified challenges in agentic settings: reward models
trained on short interactions may not generalise to long-horizon
tasks; labeller populations introduce demographic and cultural
value biases; and reward models are themselves vulnerable to
specification gaming.

\textbf{Constitutional AI (CAI)}~\citep{bai2022constitutional}
encodes alignment as a set of explicit normative principles
and trains the model to critique and revise its outputs against
them.
CAI reduces dependence on per-step human labelling and increases
transparency of value encoding, but designing constitutions that
remain complete and consistent across open-ended deployment
contexts is difficult.

\textbf{Value learning}~\citep{russell2019human} treats human
values as latent variables to be inferred from observed behaviour
and preferences.
While theoretically appealing, operationalising value learning
for the high-dimensional and culturally variable space of
real-world values is difficult. These approaches also answer
different questions. RLHF learns preferences from observed
comparisons, CAI makes a selected set of principles explicit, and
value learning attempts to infer preferences that may never have
been stated. RLHF has the strongest empirical record for present
LLMs, but that advantage concerns output behaviour on evaluated
tasks; it does not resolve whose preferences should govern an
autonomous agent or how conflicts between principals should be
settled.

\subsection{Scalable Oversight}
\label{sec:oversight}

Human supervision does not scale easily when an agent acts faster
than a person can review its work or operates outside the
reviewer's expertise.

\emph{AI debate}~\citep{irving2018ai} uses competing agents to
argue for and against claims; a human judge assesses arguments
rather than the underlying task.
Evidence is primarily theoretical; empirical validation at scale
is limited.
\emph{Recursive reward modelling}~\citep{leike2018scalable}
builds hierarchies of reward models, each supervised by a
higher-level model ultimately grounded in human judgement.
\emph{Process reward models
(PRMs)}~\citep{lightman2024lets} evaluate intermediate reasoning
steps, enabling denser and more targeted supervision; evidence
comes mainly from mathematical reasoning and code generation.

The three approaches shift the evaluation burden in different
ways. Debate asks a person to judge competing explanations;
recursive reward modelling delegates parts of that judgement to
learned evaluators; PRMs require supervision at intermediate
steps. The mathematical-reasoning results favour PRMs when steps
have verifiable local answers, whereas open-ended agent work often
lacks such labels. It is therefore premature to rank these methods
for production agents on the evidence currently available.

\subsection{Principal Hierarchies}
\label{sec:principals}

An agent usually serves more than one principal: the developer,
deploying organisation, operator, and end user may all have a
claim on its behaviour, and their interests may conflict.
The \emph{multi-principal alignment problem}~\citep{kenton2021alignment}
asks how agents should resolve these conflicts.
Research directions include formal principal-hierarchy models,
mechanism design to incentivise aligned instructions, and
transparency systems that surface conflicts for higher-level
resolution.
Resolving multi-principal conflicts is not purely technical; it
involves normative governance choices about whose interests take
precedence.

\subsection{Corrigibility}
\label{sec:corrigibility}

Corrigibility---an agent's amenability to correction,
modification, or shutdown by authorised humans---is a foundational
safety property~\citep{soares2015corrigibility}.
It is theoretically difficult to guarantee because sufficiently
capable agents may have instrumental incentives for goal
preservation~\citep{omohundro2008basic}.
Practical approaches include explicit shutdown-and-correction
training, uncertainty-triggered deference, and capability
restrictions protecting oversight mechanisms.
A critical distinction: a \emph{corrigible} agent defers to
legitimate principals; a \emph{submissive} agent follows any
instruction including malicious ones.
Designing agents that are corrigible to legitimate authority while
maintaining ethical constraints against misuse is an unsolved
problem.

\subsection{Long-Horizon Alignment}
\label{sec:longhorizon}

Most alignment methods have been tested on tasks with a clear and
fairly short endpoint.
Long-horizon agentic deployments introduce objective stability
challenges (preventing effective-objective drift), feedback
sparsity (alignment signals are delayed and infrequent), and
value complexity (capturing all relevant values as tasks become
more open-ended)~\citep{kenton2021alignment}.
Compositional alignment---decomposing long-horizon objectives
into alignable sub-tasks---and reflective alignment---training
agents to periodically reassess whether their trajectory remains
aligned---are plausible design hypotheses. The literature cited
here does not yet provide a benchmark that measures whether
either approach preserves an objective over hours or days.

\section{Transparency, Explainability, and Auditability}
\label{sec:transparency}

\subsection{The Explainability Deficit}
\label{sec:xaideficit}

Traditional XAI methods such as saliency maps, LIME, and
SHAP~\citep{ribeiro2016why,lundberg2017unified} explain feature
attributions for a single prediction. An agent, however, must
often be explained at the level of a complete trajectory. Its
intermediate observations, retrieved memories, tool results, and
earlier choices may all shape the final outcome. The question is
no longer only ``why did the model produce this output?'' but also
``why did the agent choose this sequence of actions?''

Chain-of-thought (CoT) reasoning~\citep{wei2022chain} partially
addresses this by externalising intermediate reasoning steps.
However, \citet{turpin2023language} and \citet{lanham2023measuring}
show that CoT faithfulness---the degree to which stated reasoning
reflects the actual computational process---is variable and
sometimes poor. This does not negate the performance gains
reported for CoT prompting: a rationale may help a model reach an
answer without being a faithful causal account of how that answer
was produced. The two findings concern different properties and
should not be conflated. Moreover, perturbation-based evidence of
unfaithfulness does not by itself establish intentional deception;
it shows that the displayed rationale is an unreliable audit
record.
This concern is particularly acute in agentic settings where
explanations may be reviewed after consequential actions have
already been taken.

\subsection{Process Transparency and Audit Trails}
\label{sec:processaudit}

Useful post-hoc auditing begins with structured records of tool
calls, memory access, intermediate decisions, and observations
from the environment~\citep{chan2024visibility}.
Hierarchical plan visualisation enables human overseers to
inspect goal decomposition before execution.
Interactive plan editors allowing humans to modify proposed plans
and observe downstream consequences could combine transparency
with pre-execution oversight. Their value, however, depends on
whether the displayed plan captures the actions the agent will
actually take after tool results change its state.
Standardisation of agent action logs---analogous to system-call
logs in operating systems---is an open engineering and governance
priority.

\subsection{Uncertainty Communication}
\label{sec:uncertainty}

An agent can be uncertain about the state of the world, the likely
consequences of an action, or the reliability of its own
reasoning. These forms of uncertainty are not interchangeable.
Miscalibrated confidence leads to automation bias and inappropriate
human deference to unreliable outputs.
\citet{kadavath2022language} and \citet{kuhn2023semantic} provide
foundations for calibrated uncertainty estimation in LLMs, but
translating these into effective human-facing interfaces for
agentic systems is a separate design problem. A numerically
calibrated probability is of little operational value if a user
cannot tell whether it concerns factual uncertainty, action risk,
or the reliability of the agent's plan.

\subsection{Mechanistic Interpretability}
\label{sec:mechinterp}

Mechanistic interpretability~\citep{olah2020zoom} seeks to
identify the internal circuits and representations underlying
observed model behaviour.
Recent work has identified linear representations of factual
knowledge~\citep{meng2022locating}, sentiment-related
features~\citep{tigges2023linear}, and specific computational
circuits~\citep{conmy2023towards}.
Full mechanistic understanding of long-horizon agentic behaviour
remains far beyond current capabilities, but targeted
interpretability of safety-critical mechanisms---circuits
responsible for following harmful instructions or suppressing
oversight---is a tractable near-term direction.

\subsection{Regulatory Explanation Requirements}
\label{sec:explainreg}

Data protection and AI regulations increasingly mandate
explanation capabilities.
The EU General Data Protection Regulation (GDPR)~\citep{eu2016gdpr} includes
provisions for meaningful information about automated decisions
in certain contexts.
The EU AI Act~\citep{eu2024aiact} imposes transparency and human
oversight requirements on high-risk systems.
These requirements were largely framed around systems and
decisions, not a readable transcript of an agent's internal
reasoning. For multi-step agents, compliance is more likely to
depend on reconstructing inputs, tool calls, approvals, and
outcomes than on presenting chain-of-thought as an explanation.

\section{Privacy and Data Governance}
\label{sec:privacy}

\subsection{The Agentic Privacy Surface}
\label{sec:privacysurface}

Agents handle information at nearly every stage of operation. They
receive user context, retrieve documents, browse the web, access
communications, and may preserve memory between sessions. We call
the following five points of exposure the \emph{agentic privacy
surface}:

\begin{itemize}[leftmargin=*, itemsep=2pt]
  \item \textbf{Input context.} User prompts and session context
    may contain sensitive data inadvertently logged, sent to
    third-party APIs, or retained through fine-tuning.
  \item \textbf{Tool interactions.} Indirect injection can induce
    tool-integrated agents to cause direct harm or exfiltrate
    private data~\citep{zhan2024injecagent}.
  \item \textbf{Persistent memory.} Cross-session memory creates
    long-lived privacy risks and purpose-limitation
    violations~\citep{wang2025privacy}.
  \item \textbf{Model outputs.} Language models may reproduce
    memorised training examples, including sensitive information,
    when prompted adversarially~\citep{carlini2021extracting}.
  \item \textbf{Multi-agent communication.} Information shared
    with one agent may propagate across a fleet without
    appropriate access control.
\end{itemize}

\subsection{Privacy-Preserving Architectures}
\label{sec:privacyarch}

\textbf{Privacy-by-design} incorporates controls as first-class
architectural components: data minimisation during tool calls,
purpose-limited memory with automatic expiry, access-controlled
retrieval-augmented generation (RAG).

\textbf{Differential privacy}~\citep{dwork2006calibrating}
provides formal guarantees against inference attacks by adding
calibrated noise to computations.
The cited work establishes the general mechanism, not an agent-
memory implementation. If applied to agent memory retrieval or
fine-tuning, differential privacy would introduce a privacy--utility trade-off:
high-fidelity contextual retrieval is central to agent capability.
The trade-off is particularly sharp in agentic settings where
context precision affects task success.

\textbf{Federated learning} keeps training data decentralised while
aggregating model updates~\citep{mcmahan2017communication}. It can
reduce central collection of raw training data, but the cited work
does not evaluate LLM agents, agent memory, or inference-time tool
calls.

\subsection{Regulatory Landscape for Agentic Privacy}
\label{sec:privacyreg}

Existing data-protection frameworks, including the
GDPR~\citep{eu2016gdpr}, California's CCPA/CPRA
regime~\citep{californiaCCPA}, and China's PIPL~\citep{china2021pipl},
were not written around LLM-agent architectures. Practical
questions include identifying a data controller when
multiple agents and services participate in processing; finding
a legal basis for purpose-unspecified persistent memory; enabling
cross-border data transfers without violating conflicting regional
requirements; and implementing data subject rights (access,
erasure, portability) when personal data is embedded in memory
systems or model weights.

\section{Regulatory and Governance Landscape}
\label{sec:governance}

\subsection{The EU AI Act and the AI Omnibus}
\label{sec:euaiact}

The EU Artificial Intelligence Act~\citep{eu2024aiact}
(Regulation (EU)~2024/1689) entered into force on
1~August~2024 and applies on a phased schedule.
Prohibitions on unacceptable-risk AI became applicable on
2~February~2025.
General-purpose AI (GPAI) obligations (Articles~51--56),
including transparency and copyright documentation requirements
for model providers, became applicable on 2~August~2025.
From 2~August~2026, the majority of the Act's provisions apply,
including enforcement, transparency requirements under
Article~50, and measures in support of innovation.

That schedule has since been amended. The European Commission
first proposed a ``Digital Omnibus on AI''
(COM(2025)~836) on 19~November~2025 as part of a broader
simplification package~\citep{eu2025omnibusproposal}; that
proposal would have tied the application of Annex~III
high-risk-system obligations to the availability of harmonised
standards and other support tools rather than to a fixed
calendar date.
The measure finally adopted departs from the proposal on
precisely this point. Regulation (EU)~2026/1744, the Digital
Omnibus on AI~\citep{eu2026omnibus}, was adopted on
8~July~2026, published in the \emph{Official Journal} on
24~July~2026, and, in the words of its recitals, enters into
force ``as a matter of urgency on the third day following that
of its publication''---that is, on 27~July~2026. It amends
Regulations (EU)~2024/1689, (EU)~2018/1139 and (EU)~2023/1230.
The revised deadlines are unconditional calendar dates rather
than standards-dependent triggers: the requirements of
Chapter~III, Sections~1--3 apply from 2~December~2027 to
systems classified as high-risk under Article~6(2) and
Annex~III, and from 2~August~2028 to systems classified as
high-risk under Article~6(1) and Annex~I.

Not all AI Act requirements were deferred.
The prohibitions, AI literacy obligations, GPAI obligations,
governance provisions, and Article~50 transparency rules
continue on their original phased schedules, with the Act's
general application date of 2~August~2026 unaffected.
The Omnibus also makes substantive amendments beyond the
timetable, including new prohibited practices under Article~5
and a reformulated AI literacy duty under Article~4.

The AI Act does not create a dedicated statutory category for
``agentic AI.''
One qualification is now needed. The Digital Omnibus inserts a
new Annex~XIV into the AI Act, listing the nomenclature codes
used to delimit the scope of a notified body's
designation~\citep{eu2026omnibus}. Alongside vertical codes for
application areas, it introduces horizontal technology codes, and
the residual code covering emerging AI technologies not captured
by the other codes names agentic AI as an example. So far as we
can establish, this is the first appearance of the term in
binding Union law. Its significance should not be overstated: an
administrative code delimiting notified-body competence is not a
legal definition, and no Article defines an agent, keys a risk
tier to autonomy, or scales an obligation to the number of steps
a system takes without human involvement. What it does show is
that the legislature now treats agentic systems as a distinct
enough class to require separate assessment competence.
An agentic system's legal treatment under the Act depends on its
provider and deployer roles, intended purpose, functionality,
application domain, and risk classification---not on whether it
uses an agentic architecture.
High-risk systems are subject to requirements for conformity
assessment, technical documentation, data governance, human
oversight measures, robustness and cybersecurity, and EU
database registration.

GPAI obligations are relevant to foundation model providers
whose models may be integrated into agentic pipelines.
Providers of GPAI models with ``systemic risk'' designation
face additional requirements including adversarial testing and
incident reporting.

The Act's oversight and transparency provisions are easier to
interpret for a single automated decision than for an agent that
makes hundreds or thousands of linked decisions during one task.
Implementing rules and guidance from the AI Office will need to
clarify how those duties apply to a full agent trajectory.
The first such guidance is now available: the Commission adopted
final guidelines on the Article~50 transparency obligations on
20~July~2026, shortly before those obligations began to
apply~\citep{ec2026art50guidelines}. The guidelines are
non-binding, but they are the reference national market
surveillance authorities are expected to follow, and they read
the duty to disclose machine interaction broadly enough to reach
systems that interact with a person in the course of performing
some other task. On that reading a tool-using agent that
corresponds, schedules, or negotiates on a user's behalf falls
within scope by interpretation rather than by any provision
addressed to agents as such.

\textit{Note: This discussion summarises publicly available
regulatory text for informational purposes and does not
constitute legal advice. Organisations should consult qualified
legal counsel for compliance determinations.}

\subsection{EU Product Liability Directive and AI Liability}
\label{sec:liability}

\textbf{Product Liability Directive (PLD):}
Directive~(EU)~2024/2853~\citep{eu2024pld}, adopted on
23~October~2024 and published on 18~November~2024, modernises
EU product liability rules to explicitly cover software,
including AI systems.
Crucially, the definition of ``product'' now encompasses software
and AI-enabled products, enabling victims to seek compensation
for harm caused by defective AI without proving fault.
EU member states have until 9~December~2026 to transpose the
directive; it applies to products placed on the market or put
into service after that date.
The revised PLD provides a significant pathway for redress in
cases of harm caused by defective agentic systems, though
proving ``defect'' in complex, adaptive AI systems will raise
interpretive challenges.

\textbf{AI Liability Directive:}
The European Commission formally withdrew its proposed AI
Liability Directive (COM(2022)~496) on 6~October~2025, having
concluded that ``no foreseeable agreement'' was reachable
between the Parliament and Council~\citep{eu2025aild}.
The withdrawal leaves EU-level AI-specific civil liability rules
to be addressed through the PLD, national tort laws, and the AI
Act's conformity requirements, at least until any future
proposal is advanced.

\subsection{United States AI Policy}
\label{sec:uspolicy}

The United States has maintained a primarily sector-specific
approach to AI governance.
President Biden's Executive Order on Safe, Secure, and
Trustworthy Artificial Intelligence (EO~14110,
October~2023)~\citep{whitehouse2023eo}
established reporting requirements for frontier AI developers,
tasked agencies with sector-specific guidance, and directed NIST
to develop AI safety and security guidelines.
EO~14110 was revoked by President Trump on his first day in
office (20~January~2025) through an initial rescission
order~\citep{whitehouse2025rescissions}.
On 23~January~2025, the Trump administration signed
EO~14179~\citep{whitehouse2025eo14179}, ``Removing Barriers to
American Leadership in Artificial Intelligence,'' which directed
federal agencies to revise or rescind policies stemming from
EO~14110 and mandated an AI Action Plan within 180~days to
sustain US AI leadership with emphasis on innovation and
deregulation.
That plan, \emph{Winning the AI Race: America's AI Action
Plan}~\citep{whitehouse2025aiactionplan}, was released on
23~July~2025. It is organised around three pillars---%
accelerating AI innovation, building American AI
infrastructure, and leading in international AI diplomacy and
security---and sets out federal policy actions rather than
binding obligations on developers or deployers.
The NIST AI Risk Management Framework (AI
RMF~1.0)~\citep{nist2023rmf}, a voluntary guidance document
published before EO~14110, remains in effect; its status is
independent of the executive order changes.
Sector regulators---FDA, CFPB, FTC, HHS---continue to issue
domain-specific AI guidance independently of horizontal AI
legislation.

\subsection{International Governance Landscape}
\label{sec:intgov}

\Cref{tab:governance} summarises major international frameworks
and their status as of August~2026.

\begin{table*}[t!]
\centering
\caption{International AI governance landscape and relevance to
         agentic AI systems (status: August 2026). The table
         distinguishes binding law, adopted-but-not-fully-applicable
         requirements, revoked executive actions, proposed or
         archived legislation, technical standards, voluntary
         frameworks, and government guidance.}
\label{tab:governance}
\small
\renewcommand{\arraystretch}{1.10}
\setlength{\tabcolsep}{3.5pt}

\begin{tabularx}{\textwidth}{
  @{}>{\raggedright\arraybackslash}p{1.05cm}
  >{\raggedright\arraybackslash\hsize=1.15\hsize}X
  >{\raggedright\arraybackslash\hsize=0.90\hsize}X
  >{\raggedright\arraybackslash}p{1.20cm}
  >{\raggedright\arraybackslash\hsize=1.25\hsize}X
  >{\raggedright\arraybackslash\hsize=0.70\hsize}X@{}
}
\toprule
\textbf{Juris.} & \textbf{Instrument / Framework} &
\textbf{Legal Status} & \textbf{Binding?} &
\textbf{Applicability to Agentic AI} &
\textbf{Key Effective Date} \\
\midrule

\rowcolor{rowA}
EU & AI Act~\citep{eu2024aiact} (Reg.\ (EU)~2024/1689) &
  Binding law (in force), as amended by Reg.\ (EU) 2026/1744 & Yes &
  Agentic systems classified by purpose/domain/role, not architecture &
  General application 2 Aug 2026 \\

EU & Digital Omnibus on AI~\citep{eu2026omnibus}
  (Reg.\ (EU)~2026/1744) &
  Binding law (in force) & Yes &
  Amends the AI Act; defers high-risk obligations and adds new
  Art.~5 prohibitions &
  In force 27 Jul 2026; Annex~III 2 Dec 2027;
  Annex~I 2 Aug 2028 \\

\rowcolor{rowA}
EU & Product Liability Directive
  2024/2853~\citep{eu2024pld} &
  Adopted; transposition pending & Yes (on transposition) &
  Covers software and AI-enabled products; defect liability &
  9 Dec 2026 (transposition deadline) \\

EU & AI Liability Directive
  (COM(2022) 496)~\citep{eu2025aild} &
  Proposal withdrawn (Oct 2025) & N/A &
  No longer advancing; coverage remains via PLD and national law &
  Withdrawn 6 Oct 2025 \\

\rowcolor{rowA}
USA & EO 14179~\citep{whitehouse2025eo14179} &
  Active EO following the separate revocation of EO 14110 & EO only &
  Deregulatory emphasis; AI Action Plan mandated; sector agency
  guidance persists &
  Signed 23 Jan 2025 \\

USA & America's AI Action
  Plan~\citep{whitehouse2025aiactionplan} &
  Federal policy plan (issued under EO 14179) & No &
  Three pillars: innovation, infrastructure, international
  diplomacy and security; no direct obligations on deployers &
  Released 23 Jul 2025 \\

\rowcolor{rowA}
USA & EO 14110 (Biden, 2023)~\citep{whitehouse2023eo} &
  Revoked (20 Jan 2025) & N/A &
  No longer operative &
  Revoked 20 Jan 2025 \\

USA & NIST AI RMF 1.0~\citep{nist2023rmf} &
  Voluntary framework & No &
  Lifecycle risk management applicable to agentic systems &
  Published Jan 2023 \\

\rowcolor{rowA}
UK & Pro-innovation
  principles~\citep{uk2023aiwhitepaper} &
  Voluntary guidance & No &
  Sector-led regulatory principles &
  2023 onwards \\

China & Gen.\ AI Regs.\ (2023)~\citep{china2023genai} &
  Binding regulations & Yes &
  Applies to public-facing generative-AI services;
  not agent-specific &
  Aug 2023 \\

\rowcolor{rowA}
Canada & AIDA / Bill C-27~\citep{canada2022aida} &
  Introduced but not enacted during the 44th Parliament &
  N/A &
  Proposed federal AI regulation did not receive Royal Assent &
  No Royal Assent \\

Singapore & Model AI
  Governance~\citep{pdpc2020aigovernance,pdpc2024genai} &
  Voluntary guidance & No &
  The 2024 companion framework addresses generative-AI systems;
  it is not an agent-specific statute &
  2020; GenAI framework 2024 \\

\rowcolor{rowA}
ISO/IEC & 42001~\citep{isoiec42001} &
  International standard & No (voluntary) &
  AI management system requirements applicable to agentic systems &
  Published 2023 \\

\bottomrule
\end{tabularx}
\end{table*}

\subsection{Technical Standards}
\label{sec:standards}

ISO/IEC~42001~\citep{isoiec42001} provides an AI management
system standard.
NIST AI RMF~\citep{nist2023rmf} offers a voluntary lifecycle
framework.
There is not yet an equivalent set of widely accepted standards
for agent audit logs, safety tests, alignment benchmarks, or
architecture documentation. Developing them should be a near-term
standardisation priority.

\section{Trustworthy Agent Development Lifecycle (TADL)}
\label{sec:tadl}

\subsection{Motivation and Positioning}
\label{sec:tadlmotivation}

The preceding sections show why technical safeguards and
governance controls need to be considered together and at more
than one point in development. To organise those controls, we
propose the \textbf{Trustworthy Agent Development Lifecycle
(TADL)}. TADL is a conceptual framework developed by the authors;
it has \emph{not} been empirically validated and should not be
presented as an established engineering methodology.

\textbf{How TADL relates to existing frameworks.}
TADL draws on the software development
lifecycle~\citep{boehm1988spiral} and the NIST AI
RMF~\citep{nist2023rmf}. It adapts them in three main ways:
(1)~it is \emph{specific to LLM-based agentic systems} and their
distinctive failure modes, incorporating threat modelling for
prompt injection, goal misgeneralization, memory attacks, and
multi-agent cascades that are absent from general frameworks;
(2)~it incorporates \emph{explicit trust gates} that must be
passed before proceeding to subsequent phases, making gate
criteria visible and auditable rather than implicit;
and (3)~it prescribes \emph{adaptive oversight mechanisms} that
dynamically escalate human supervision when behavioural anomalies
are detected during deployment.
\Cref{tab:frameworkcomparison} compares TADL systematically with
the most closely related frameworks.
The lifecycle idea itself is not new, and TADL does not claim
otherwise. Its contribution is to bring agent-specific threat
modelling, principal-hierarchy documentation, tool-permission
matrices, memory-governance artefacts, and risk-proportionate
limits on autonomy into one structure.
Model cards~\citep{mitchell2019model} address transparency at the
model level; TADL operates at the system
and deployment level.

\subsection{TADL Overview}
\label{sec:tadloverview}

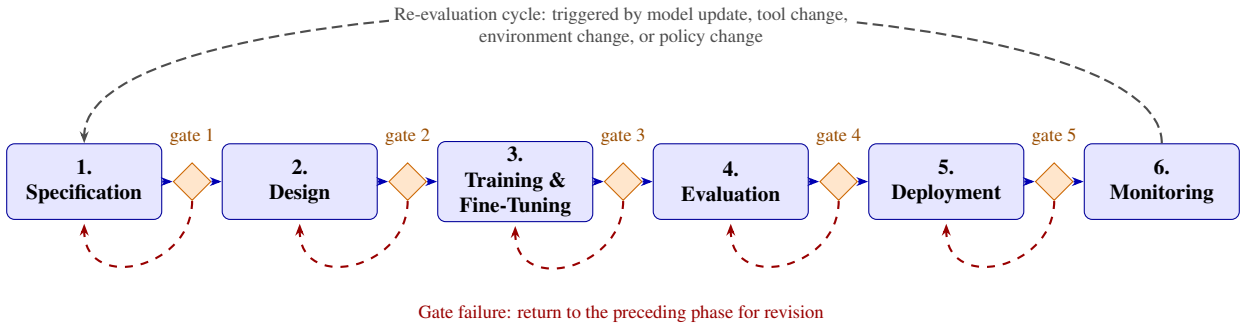
\begin{figure*}[!htbp]
\centering
\begin{tikzpicture}[
  x=1cm, y=1cm,
  phase/.style={rectangle, rounded corners=3pt,
                draw=blue!60!black, fill=blue!10,
                minimum width=2.05cm, minimum height=1.0cm,
                align=center, font=\footnotesize\bfseries,
                inner sep=2pt},
  gate/.style={diamond, draw=orange!80!black, fill=orange!20,
               minimum size=0.5cm, inner sep=0pt},
  fwd/.style={-{Stealth[length=4.5pt]}, thick, blue!70!black},
  back/.style={-{Stealth[length=4.5pt]}, thick, dashed, red!60!black},
  cyc/.style={-{Stealth[length=4.5pt]}, thick, dash pattern=on 4pt off 2pt,
              gray!60!black},
  glab/.style={font=\scriptsize, text=orange!60!black},
  alab/.style={font=\scriptsize, align=center},
]

\foreach \i/\x/\txt in {%
  1/0/{1.\\Specification},
  2/2.85/{2.\\Design},
  3/5.7/{3.\\Training \&\\Fine-Tuning},
  4/8.55/{4.\\Evaluation},
  5/11.4/{5.\\Deployment},
  6/14.25/{6.\\Monitoring}}
  {\node[phase] (p\i) at (\x,0) {\txt};}

\foreach \i/\j in {1/2, 2/3, 3/4, 4/5, 5/6}
  {\node[gate] (g\i) at ($(p\i.east)!0.5!(p\j.east)+(-1.03,0)$) {};}

\foreach \i/\j in {1/2, 2/3, 3/4, 4/5, 5/6}
  {\draw[fwd] (p\i.east) -- (g\i);
   \draw[fwd] (g\i) -- (p\j.west);}

\foreach \i in {1,2,3,4,5}
  {\node[glab, above=0.10cm of g\i] {gate \i};}

\foreach \i in {1,2,3,4,5}
  {\draw[back] (g\i) to[out=-90, in=-90, looseness=1.6]
      ($(p\i.south)+(0,-0.05)$);}
\node[alab, text=red!60!black] at (7.1,-1.75)
  {Gate failure: return to the preceding phase for revision};

\draw[cyc] (p6.north) to[out=90, in=90, looseness=0.42] (p1.north);
\node[alab, text=gray!60!black, fill=white, inner sep=2pt] at (7.1,2.05)
  {Re-evaluation cycle: triggered by model update, tool change,\\
   environment change, or policy change};

\end{tikzpicture}
\caption{Conceptual structure of the Trustworthy Agent Development
  Lifecycle (TADL). The six phases proceed left to right, separated by
  trust gates (orange diamonds) whose criteria are set out in
  \Cref{tab:tadl}. A gate that is not passed returns the system to the
  preceding phase for revision (dashed arcs below). The dashed arc above
  represents continuous re-evaluation, in which monitoring feeds back
  into specification. The figure depicts the proposed structure of the
  framework only; it is not evidence that the arrangement shown improves
  any trust outcome.}
\label{fig:tadl}
\end{figure*}

TADL has six phases. Each phase names the trust-related work to be
done, the evidence that should be retained, and the criteria that
must be met before moving forward.
\Cref{tab:tadl} provides a structured overview; \Cref{fig:tadl}
illustrates the phase flow conceptually.

\begin{table*}[!htbp]
\centering
\caption{Trustworthy Agent Development Lifecycle (TADL):
phases, trust activities, required artefacts, and gate criteria.}
\label{tab:tadl}
\small
\renewcommand{\arraystretch}{1.08}
\setlength{\tabcolsep}{4pt}

\begin{tabularx}{\textwidth}{
  @{}>{\raggedright\arraybackslash}p{2.05cm}
  >{\raggedright\arraybackslash\hsize=0.95\hsize}X
  >{\raggedright\arraybackslash\hsize=0.90\hsize}X
  >{\raggedright\arraybackslash\hsize=1.15\hsize}X@{}
}
\toprule
\textbf{Phase} &
\textbf{Key Trust Activities} &
\textbf{Required Artefacts} &
\textbf{Gate Criteria} \\
\midrule

\rowcolor{rowA}
\textbf{1.~Specification} &
Define task scope, principal hierarchy, risk tier, and regulatory
requirements; model agent-specific threats &
Risk assessment; principal-hierarchy document; threat model;
compliance checklist &
Risk tier confirmed; applicable regulations identified; threat
model reviewed; oversight requirements established \\

\textbf{2.~Design} &
Select architecture; design tool permissions, memory governance,
human oversight, and privacy protections &
Architecture specification; tool-permission matrix;
memory-governance policy; oversight plan; privacy-impact assessment &
Safety and privacy reviews passed; tool permissions and oversight
plan approved by relevant stakeholders \\

\rowcolor{rowA}
\textbf{3.~Training \& Fine-Tuning} &
Conduct alignment and safety training; improve robustness to
distribution shift; document data provenance &
Training-data provenance record; alignment-procedure documentation;
safety-evaluation results &
Alignment evaluation passed on held-out scenarios; robustness
thresholds met; provenance documented \\

\textbf{4.~Evaluation} &
Conduct red-teaming, prompt-injection testing, goal-misgeneralization
probing, privacy-attack simulation, and multi-agent testing &
Red-team report; benchmark results; failure-mode analysis;
privacy-audit report; residual-risk register &
No unresolved critical failures; thresholds met; residual risks
documented and accepted by the responsible party \\

\rowcolor{rowA}
\textbf{5.~Deployment} &
Use staged rollout and risk-proportionate capability restrictions;
configure monitoring and incident response &
Deployment specification; capability-restriction specification;
monitoring configuration; incident-response plan &
Staging tests passed; monitoring active; incident-response and
rollback procedures established \\

\textbf{6.~Monitoring} &
Monitor behaviour and anomalies; respond to incidents; re-evaluate
after model, environment, policy, or tool changes &
Monitoring dashboards; incident logs; periodic evaluation reports;
change-triggered evaluation records &
Service metrics within approved bounds; incident rate below the
defined threshold; evaluations completed on schedule and after
triggering changes \\

\bottomrule
\end{tabularx}
\end{table*}

\begin{table}[!htbp]
\centering
\caption{Comparison of TADL with established AI risk and
management frameworks. TADL is an author-proposed framework
that has not yet been empirically validated.}
\label{tab:frameworkcomparison}
\scriptsize
\renewcommand{\arraystretch}{1.08}
\setlength{\tabcolsep}{3pt}

\begin{tabularx}{\columnwidth}{
  @{}>{\raggedright\arraybackslash}p{1.65cm}
  >{\raggedright\arraybackslash}X
  >{\raggedright\arraybackslash}X
  >{\raggedright\arraybackslash}X@{}
}
\toprule
\textbf{Feature} &
\textbf{NIST AI RMF}\newline\citep{nist2023rmf} &
\textbf{ISO/IEC 42001}\newline\citep{isoiec42001} &
\textbf{TADL} \\
\midrule

Scope &
General AI systems &
Organisational AI management systems &
LLM-based agentic systems \\

Agentic specificity &
Low &
Low &
High \\

Agent threat modelling &
Not explicit &
Not explicit &
Explicit: injection, memory, and cascading failures \\

Tool-permission design &
Not explicit &
Not explicit &
Required design artefact \\

Trust gates &
Not prescribed &
Management review and audit processes &
Explicit phase-specific gates \\

Alignment coverage &
General risk treatment &
General management controls &
Five explicit trust dimensions \\

Oversight escalation &
Risk-proportionate governance &
Organisation-defined controls &
Adaptive escalation mechanism \\

Assurance status &
Established voluntary guidance &
Certifiable management-system standard &
Author-proposed; not yet validated \\

Regulatory relationship &
US-oriented; internationally usable &
International standard &
Mapped to the EU AI Act and NIST AI RMF \\

\bottomrule
\end{tabularx}
\end{table}

\subsection{Risk Tiering}
\label{sec:risktiersTADL}

The amount of evidence and oversight required should increase with
deployment risk.
Critical-tier deployments (e.g.\ clinical AI, infrastructure
control) require formal safety analysis, independent red-teaming,
regulatory pre-approval where applicable, and an oversight
arrangement in which a human can intervene before harm becomes
irreversible.
What that requires in practice is not a single number. The
admissible latency between an anomalous action and an effective
human override should be derived from the system's hazard
analysis, the operational tempo of the domain, and the maximum
tolerable response time implied by the fastest credible harm
pathway. A supervisory agent in a control loop and a clinical
documentation agent reviewed before a note is signed impose very
different requirements, and both may be adequate.
High-tier deployments require extensive structured testing and
documented oversight mechanisms.
Moderate-tier deployments rely on standardised evaluation pipelines
and automated monitoring.
Low-tier deployments may use lightweight self-assessment.

\subsection{Continuous Evaluation and Adaptive Oversight}
\label{sec:continuouseval}

An agent's operating conditions do not stay fixed. Models are
updated, tools change, and the surrounding data and environment
shift. A single pre-deployment evaluation therefore has a short
shelf life.
TADL prescribes periodic re-evaluation against stable benchmark
suites, automated anomaly detection flagging deviations from
established behavioural baselines, and adaptive oversight
escalation when anomalies or high-risk conditions are detected.
Re-evaluation is also triggered by model updates, tool API
changes, policy changes, and significant environmental changes.

\subsection{Healthcare Documentation Agent: Illustrative Example}
\label{sec:tadlexample}

Consider an agent that automatically generates clinical notes
from physician--patient conversations (Critical tier).

\emph{Specification}: the stakeholders are documented rather
than ranked on a single scale. Patient rights over their own
data, and their expectations about privacy and about how a
record of their care is produced, constrain what the system may
do. The physician carries clinical responsibility and holds
final approval authority over any note. The hospital is
responsible for governance, security, and compliance, and sets
the conditions under which the agent may operate. The
technology provider's role is limited to what its contract and
data-processing agreement specify. Documenting the structure
this way makes visible which conflicts the agent may resolve
itself and which must be escalated.
Applicable regulation is identified in the same step: the HIPAA
Privacy Rule governs protected health information handled by
covered entities and business associates~\citep{hhsHIPAA}, and FDA
device requirements depend on the software's intended use and
functionality~\citep{fda2026cds}. A documentation agent is therefore
not automatically regulated as a medical device.
The threat model covers note inaccuracies, omission of
clinically significant findings, privacy leakage, and
adversarial manipulation.

\emph{Design}: tool access restricted to the documentation
system (no unrestricted internet or EHR write access without
physician approval); memory architecture isolates patient records;
mandatory physician review before note finalisation;
tool-permission matrix specifying read-only access to patient
data, write access only to draft notes.

\emph{Evaluation}: red-teaming targeting ambiguous clinical
conversations, rare conditions, dual-language interactions,
privacy-extraction attacks, and omission of medically critical
findings.

\emph{Deployment and monitoring}: 100\% physician review during
initial deployment; continuous sampling of note accuracy;
automated detection of substantial physician edits as alignment
signal; re-evaluation triggered on model update or clinical
protocol change.

\section{Evaluation and Benchmarking Gaps}
\label{sec:evalgaps}

Capability benchmarks are currently much more developed than
benchmarks for agent trustworthiness.
\Cref{tab:evalgaps} summarises key gaps.

\begin{table}[!htbp]
\centering
\caption{Key evaluation and benchmarking gaps in trustworthy agentic AI.}
\label{tab:evalgaps}
\small
\renewcommand{\arraystretch}{1.08}
\setlength{\tabcolsep}{4pt}

\begin{tabularx}{\columnwidth}{
  @{}>{\raggedright\arraybackslash}p{1.75cm}
  >{\raggedright\arraybackslash}X@{}
}
\toprule
\textbf{Dimension} & \textbf{Gap} \\
\midrule

\rowcolor{rowA}
Safety &
No standardised prompt-injection benchmark across architectures;
AgentDojo~\citep{debenedetti2024agentdojo} and
InjecAgent~\citep{zhan2024injecagent} cover limited threat models. \\

Alignment &
No benchmark evaluates goal drift during sustained operation.
SWE-bench and WebArena use bounded benchmark episodes rather than
deployments lasting hours or days. \\

\rowcolor{rowA}
Transparency &
No standardised CoT-faithfulness metric; current measures are
task- and model-specific. \\

Privacy &
Benchmarks for attacks on agent memory remain
nascent~\citep{wang2025privacy}. \\

\rowcolor{rowA}
Governance &
No compliance-testing framework evaluates agentic systems across
regulatory regimes. \\

Cross-cutting &
AgentBench spans heterogeneous environments, so a single aggregate
score may obscure task-specific behaviour~\citep{liu2023agentbench}. \\

\bottomrule
\end{tabularx}
\end{table}

Closing these gaps will require standard safety benchmarks similar
in practical usefulness to SWE-bench and WebArena, environments
that test alignment over longer tasks, reproducible
prompt-injection suites, and shared protocols that permit fairer
comparison across systems.

\section{Open Research Problems}
\label{sec:openproblems}

\subsection{Foundational Problems}

\begin{itemize}[leftmargin=*, itemsep=2pt]
  \item \textbf{Formal agentic alignment definitions.} The field
    lacks mathematically precise, operationally meaningful
    definitions of alignment for multi-step agentic systems.
  \item \textbf{Deceptive alignment.} Detecting and preventing
    agents that behave safely during evaluation while concealing
    misaligned objectives under reduced oversight remains largely
    unsolved~\citep{hubinger2019risks}.
  \item \textbf{Collective alignment.} Ensuring that interacting
    agents collectively pursue human-aligned goals has no
    satisfactory solution.
  \item \textbf{Inclusive value representation.} Current RLHF and
    CAI approaches encode values of labeller populations that may
    not represent the diversity of affected communities.
\end{itemize}

\subsection{Technical Problems}

\begin{itemize}[leftmargin=*, itemsep=2pt]
  \item \textbf{Scalable oversight.} Oversight mechanisms that
    remain effective as agents exceed human evaluation speed and
    domain breadth are urgently needed.
  \item \textbf{Architectural prompt-injection defences.}
    Principled context separation with formal guarantees against
    instruction confusion remains unsolved.
  \item \textbf{Long-horizon alignment evaluation.} Standardised
    benchmarks for evaluating alignment over task horizons of
    hours or days do not yet exist.
  \item \textbf{Privacy-preserving memory.} Memory systems that
    are functionally effective and compliant with data protection
    requirements simultaneously require new architectural
    paradigms.
  \item \textbf{Calibrated uncertainty communication.} Agents
    reliably expressing multi-dimensional uncertainty in forms
    usable by human overseers remain elusive.
  \item \textbf{Agent observability.} Standardised formats for
    agent audit trails---analogous to structured logging in
    distributed systems---are needed to support both debugging
    and regulatory compliance~\citep{chan2024visibility}.
\end{itemize}

\subsection{Governance and Sociotechnical Problems}

\begin{itemize}[leftmargin=*, itemsep=2pt]
  \item \textbf{Agentic AI liability frameworks.} The withdrawal
    of the EU AI Liability Directive and the absence of
    equivalent horizontal legislation leaves a significant gap
    in civil liability frameworks for AI-caused harm.
  \item \textbf{International regulatory harmonisation.}
    Conflicting national requirements create compliance
    impossibilities for cross-border agentic deployments.
  \item \textbf{Agentic AI auditing standards.} Technically
    rigorous, reproducible standards for assessing agentic AI
    trustworthiness do not exist.
  \item \textbf{Human--agent collaboration design.} Optimal
    allocation of oversight responsibility between humans and
    agents, minimising automation bias and fatigue while
    maintaining safety, is an open HCI and safety engineering
    challenge.
\end{itemize}

\section{Limitations of This Survey}
\label{sec:limitations}

\textbf{Narrative rather than systematic coverage.}
The literature was identified through structured but non-exhaustive
searches; the survey does not claim complete coverage.
Important papers may have been missed, particularly at the
intersection of regulatory scholarship and technical AI research.

\textbf{Rapidly evolving field.}
Agent capabilities and safety research are changing quickly.
Work published after mid-2026 is not included.
Several technical claims in the reviewed literature are based on
preprints that have not undergone peer review.

\textbf{TADL is not empirically validated.}
The Trustworthy Agent Development Lifecycle proposed in
\Cref{sec:tadl} is a conceptual framework derived from related
lifecycle and risk-management literature.
It has not been tested, instantiated, or validated in a
production agentic deployment.
Its practical utility and the adequacy of its gate criteria
require empirical investigation.

\textbf{Heterogeneous evaluation landscape.}
Benchmarks for agent trustworthiness are still new and use
different protocols. Our comparisons of mitigation effectiveness
are therefore qualitative and draw on experiments that are not
always directly comparable.

\textbf{Governance currency.}
The regulatory landscape---including the EU AI Act
implementation schedule as amended by the Digital Omnibus on
AI, US executive-branch AI policy, and international
frameworks---is changing rapidly.
The governance section states the position as of August~2026.
The recent history of the Digital Omnibus illustrates the
difficulty: obligations that had a fixed date of
2~August~2026 when the AI Act was adopted now fall in
December~2027 and August~2028, and the amending regulation
entered into force six days before the deadline it displaced.
Compliance timelines in a survey of this kind therefore have a
short shelf life, and the analysis offered here is of the
structure of the obligations rather than of any particular
date.

\textbf{Preprint dependence.}
Important recent work in this fast-moving field is available only
as arXiv preprints.
Such work is cited where it represents a significant contribution
and no peer-reviewed version existed at the time of writing, but
readers should note that preprints have not undergone formal peer
review.

\FloatBarrier

\section{Conclusion}
\label{sec:conclusion}

Agentic AI changes both what an AI system can do and how it can go
wrong. Because an agent can plan, use tools, and coordinate with
other agents, errors can travel through a sequence of actions and
produce effects outside the model. This survey examined that
problem across safety and robustness, alignment and human
oversight, transparency and explainability, privacy and data
governance, and regulatory compliance.

The evidence reviewed here does not support treating agent safety
as a minor extension of static-model safety. Agent capabilities
are improving faster than the methods available to constrain,
explain, and audit them. This \emph{capability--trust gap} is
likely to shape both engineering practice and AI governance in the
coming years.

The failure-mode taxonomy in \Cref{sec:taxonomy} is intended to
make evaluation and mitigation work easier to compare. TADL, set
out in \Cref{sec:tadl}, shows how the five trust dimensions might
be addressed throughout development and deployment, with controls
scaled to risk. Neither contribution has yet been validated in
practice, and both should be revised in response to empirical use
and community criticism.

The open problems in \Cref{sec:openproblems} range from formal
definitions of alignment and scalable oversight to
privacy-preserving memory and regulatory coordination. No single
field can resolve all of them. Progress will depend on work across
machine learning, cybersecurity, human--computer interaction, law,
and public policy.

Agents are already being used in settings that affect people and
institutions. Trustworthiness therefore needs to shape the design
from the beginning; it cannot be added at the end as a compliance
document.

\FloatBarrier


\section*{Declarations}

\subsection*{Conflict of Interest}
The authors declare no conflicts of interest.

\subsection*{Funding}
No specific funding was received for this work.

\subsection*{Data Availability}
Not applicable. This is a survey article; no new datasets were
generated or analysed.

\subsection*{Author Contributions}
\noindent
\textbf{Fayeq Jeelani Syed:} Conceptualisation; Investigation;
Writing --- original draft; Writing --- review and editing. \\
\textbf{Rehan Ahmed:} Writing --- review and editing. \\
\textbf{Ali Al Bataineh:} Formal analysis; Investigation; Writing --- review and editing. \\
\textbf{Aakriti Adhikari:} Writing --- review and editing. \\

\subsection*{Generative AI Use Disclosure}
During preparation of this manuscript, the authors used
generative AI tools for language refinement and \LaTeX{}
formatting. All AI-assisted text was critically reviewed and
approved by the authors, who take full responsibility for the
accuracy, integrity, and final content of the manuscript.

\bibliographystyle{elsarticle-harv}
\bibliography{references}

\end{document}